\documentclass[runningheads]{llncs}

\usepackage{eccv}

\usepackage{eccvabbrv}

\usepackage{url}
\usepackage[utf8]{inputenc} %
\usepackage[T1]{fontenc}    %
\usepackage{booktabs}       %
\usepackage{amsfonts}       %
\usepackage{nicefrac}       %
\usepackage{microtype}      %
\usepackage{xcolor}         %
\usepackage{makecell} 
\usepackage{graphicx} 
\definecolor{ourscolor}{HTML}{c2d1e5}
\usepackage{colortbl}  %
\usepackage{threeparttable}
\usepackage{gensymb}
\usepackage{enumitem}
\definecolor{dgreen}{RGB}{20,139,101}
\usepackage{wrapfig}
\usepackage{lipsum}
\usepackage{tcolorbox}
\tcbuselibrary{breakable}
\usepackage{fontawesome}
\usepackage{marvosym}
\usepackage{multirow}

\usepackage{bbding}
\usepackage{pifont}
\usepackage[linesnumbered,ruled]{algorithm2e}
\usepackage{tabularx}
\usepackage{array}      %
\usepackage{float}

\usepackage[accsupp]{axessibility}  %

\def\ie{\textit{i.e.}}
\def\eg{\textit{e.g.}}

\newcommand{\ryn}{\textcolor[rgb]{0.0,0.,0}}
\newcommand{\rynq}{\textcolor[rgb]{0,0,0}}
\newcommand{\fang}[1]{\textcolor[rgb]{.0,.0,.0}{#1}}

\usepackage[pagebackref,breaklinks,colorlinks,linkcolor=eccvblue,citecolor=eccvblue,urlcolor=eccvblue]{hyperref}

\usepackage{orcidlink}

\begin{document}

\def\lastandname{,}

\title{GuideMe: Multi-Domain Task Guidance and Intervention in Streaming Video}

\titlerunning{GuideMe}

\author{Fang Liu\textsuperscript{\tiny *}\inst{1}\orcidlink{0000-0002-5763-0172} \and
Jinpeng Chen\textsuperscript{\tiny *}\inst{2}\orcidlink{0000-0002-0469-4463} \and
Ke Xu\inst{3,1}\orcidlink{0000-0001-5855-3810} \and   \\
Yuhao Liu\inst{1}\orcidlink{0000-0003-0550-4788} \and 
Huankang Guan\inst{2}\orcidlink{0000-0003-0825-8658} \and
Xudong Lu\inst{4}\orcidlink{0009-0007-1699-6286} \and
Yang Bo\inst{2}\orcidlink{0009-0001-0948-9901} \and   \\
Gerhard Hancke\inst{1}\orcidlink{0000-0002-2388-3542} \and
Rui Liu\textsuperscript{\tiny \faEnvelopeO}\inst{2}\orcidlink{0000-0003-2115-8491} \and
Rynson W.H. Lau\textsuperscript{\tiny \faEnvelopeO}\inst{1,5}\orcidlink{0000-0002-8957-8129}}

\authorrunning{F.~Liu et al.}

\institute{$^{1}$ City University of Hong Kong \qquad
$^{2}$ Huawei Research\\
$^{3}$ University of Science and Technology of China \\
\qquad $^{4}$ Chinese University of Hong Kong \\
$^{5}$ City University of Hong Kong (Dongguan)}

\maketitle

\begingroup
\renewcommand{\thefootnote}{}
\footnotetext{\textsuperscript{\tiny *}~Joint first authors.\\
\textsuperscript{\tiny \faEnvelopeO}~Corresponding authors.}
\endgroup

\begin{abstract}

While multimodal Large Language Models (MLLMs) excel at offline video understanding, an interesting question of {\it how far they are from serving as a real-time procedural coach} remains unknown. 
Such a role typically requires an MLLM to continuously monitor the execution, detect mistakes, and provide corrective guidance in a closed-loop interaction.
In this paper, we construct \textbf{GuideMe}, the first multi-domain benchmark for streaming video that supports training and evaluation of MLLMs for closed-loop interactive task guidance.
It comprises 2,458 videos spanning 223.7 hours across diverse domains (\eg, cooking, object manipulation, daily-life guidance, and fitness), with 47,775 interaction samples covering next-step instructions, completion feedback, error detection, and corrective guidance.
To evaluate existing models on GuideMe, we design a three-component assessment framework to measure the capabilities of representative MLLMs, which consists of temporal-semantic bipartite matching for sequence-level alignment, behavioral classification for intervention timing, and LLM-as-a-Judge for content quality.
Extensive experiments
highlight a critical performance asymmetry: \textit{despite excelling at providing instructions, existing MLLMs consistently fail to identify execution errors and respond with corrective feedback}.
Code and data are released at \href{https://fawnliu.github.io/project/guideme/}{\faGlobe homepage}.


\end{abstract}

\section{Introduction}
\label{sec:intro}

Recent advances in Multimodal Large Language Models (MLLMs) have demonstrated strong performance on a wide range of visual understanding tasks. Representative commercial models include Gemini 3.1 Pro~\cite{gemini3.1pro2025} and GPT-4o~\cite{hurst2024gpt}, while open-source models like Qwen3-VL~\cite{bai2025qwen3vltechnicalreport}
have also shown competitive capabilities. 
Unlocking their potential for real-time interaction naturally leads to the
\emph{procedural task guidance}: enabling an AI assistant to act as a real-time coach for multi-step activities, such as assembling furniture, following a recipe, or performing a workout, by providing instructions, monitoring execution, and intervening when errors occur.

\begin{figure*}[t]
\centering
\includegraphics[width=\linewidth]{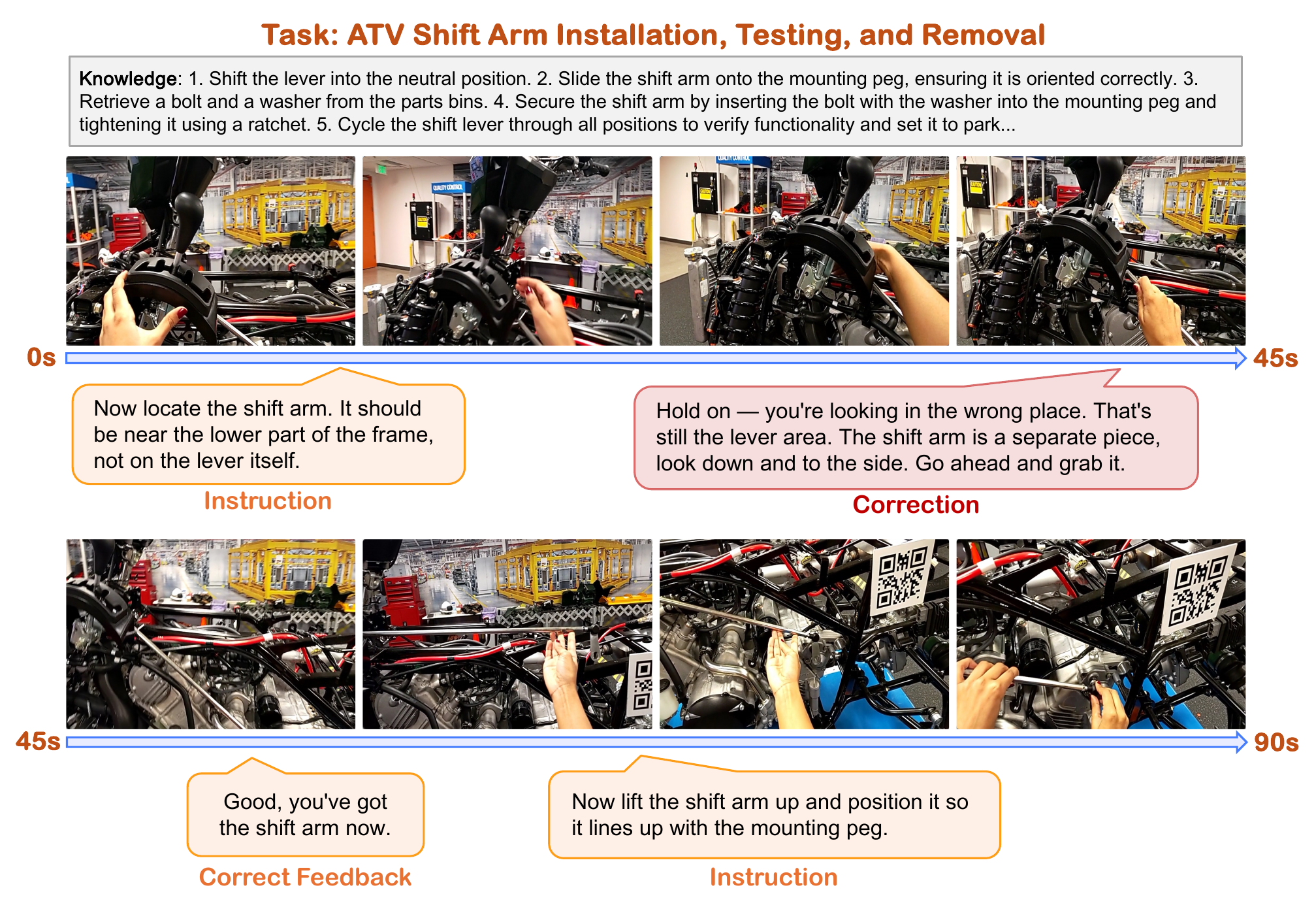}
\caption{\textbf{Data examples in GuideMe.} The benchmark captures the full closed-loop interaction cycle: the assistant issues a step-level instruction, monitors the user's execution, and upon detecting an error, provides corrective guidance to steer the user back on track. This cycle repeats throughout the procedure, enabling rigorous evaluation of interactive coaching ability.
}
\label{fig:example_guideme}
\end{figure*}

However, serving as a reliable procedural coach requires far more than understanding a video offline. In real-world interactions, users frequently deviate from the intended procedure: they skip steps, use wrong tools, or perform actions in the wrong order. An effective assistant must therefore operate in a \emph{closed loop}: continuously observing the user's actions, detecting mistakes in real time, and providing corrective feedback that guides the user back to the correct trajectory. This cycle of instructing, observing, and correcting \ryn{steps} is the defining challenge that separates a passive narrator from an active coach. 
As shown in Tab.~\ref{tab:dataset_comparison_new}, existing procedural video datasets provide step-level annotations~\cite{miech2019howto100m, damen2020epic, tang2019coin, zhou2018towards_youcook2} or post-hoc error labels~\cite{sener2022assembly101, wang2023holoassist, panchal2024say_QEVD, peddi2024captaincook4d, egoper_cvpr24}, but none of them can evaluate this complete closed-loop cycle under realistic streaming constraints.

To address these gaps, we introduce \textbf{GuideMe}, the first multi-domain streaming benchmark designed to evaluate the closed-loop corrective capabilities of AI assistants in procedural tasks. GuideMe comprises 2,458 videos spanning 223.7 hours across four diverse domains (cooking, object manipulation, daily-life guidance, and fitness), yielding a total of 47,775 interaction samples. Each sample captures one of four assistant behaviors: issuing a next-step directive, confirming correct execution, detecting an error, or providing corrective guidance. This structured interaction format enables systematic evaluation of each stage in the closed loop, from basic instruction delivery to the more challenging tasks of error detection and correction.

GuideMe is constructed through a three-stage automated annotation pipeline. 
First, we extract ordered instructional activities from fine-grained action annotations, categorizing each as a correct action, a wrong action, or a \rynq{correction action}. 
Second, we generate task knowledge, including task descriptions and canonical step sequences, from the extracted activities via LLM prompting. 
Third, we generate natural conversational dialogues aligned with action boundaries and timestamps. 
To evaluate models under streaming conditions, 
we propose a three-component assessment framework: temporal-semantic bipartite matching for sequence-level alignment, behavioral classification for instance-level intervention timing, and LLM-as-a-Judge for content quality.

Experiments across proprietary models (\eg, Gemini~\cite{gemini3.1pro2025}, Doubao~\cite{seed2025vision, seed2026vision}) and open-source models (\eg, Qwen3-VL~\cite{bai2025qwen3vltechnicalreport}, MMDuet2~\cite{wang2025mmduet2}) reveal a striking asymmetry in closed-loop performance: models achieve reasonable scores on basic instruction delivery, but their performance degrades sharply on error detection and corrective guidance, which are fundamental for reliable interactive coaching. This finding exposes a fundamental limitation: \emph{while current MLLMs can describe what should happen, they are far from coaching a user based on what is actually happening}. Our contributions are summarized as follows:
\begin{itemize}
\item We introduce GuideMe, the first multi-domain streaming benchmark for interactive procedural guidance. Unlike prior work restricted to offline annotations or single domains, GuideMe unifies four diverse domains (cooking, object manipulation, daily-life guidance, and fitness) into a closed-loop evaluation framework that covers instructions, execution feedback, error detection, and corrective guidance.

\item We develop a scalable three-stage annotation pipeline that transforms heterogeneous procedural datasets into unified streaming interaction samples. 
We 
propose a three-component evaluation framework consisting of temporal-semantic bipartite matching, behavioral classification, and LLM-as-a-Judge for rigorous assessment under streaming conditions.

\item We conduct comprehensive experiments across a wide range of models and uncover a critical capability gap. While current MLLMs can deliver step-by-step instructions, they fail at the harder stages of the closed loop, particularly detecting execution errors and providing corrective feedback, highlighting a clear direction for future research on interactive AI assistants.

\end{itemize}

\begin{table*}[t]
\centering
\small
\setlength{\tabcolsep}{2pt}
\caption{
\textbf{Comparison of procedural video datasets.} GuideMe is the only benchmark that supports all five evaluation dimensions, including the critical \textit{closed-loop interaction} where the assistant's corrective guidance modifies the user's subsequent actions within the same session.
}
\label{tab:dataset_comparison_new}
\resizebox{0.98\textwidth}{!}{
\begin{tabular}{l c c c c c c c c}
\toprule
Dataset & Domain & \#Videos & Hours 
& \makecell[c]{Step-level\\Instruction} 
& \makecell[c]{Timed\\Feedback} 
& \makecell[c]{Error\\Detection} 
& \makecell[c]{Action-specific\\Correction} 
& \makecell[c]{Closed-loop\\Interaction} \\
\midrule
Epic-Kitchen-100~\cite{damen2020epic}  &Cooking  & 700 &100  &\checkmark &$\times$ &$\times$ &$\times$ &$\times$  \\
EpicTent~\cite{jang2019epic_tent}    &Tent Making  &7  &5.4   & \checkmark  &$\times$  &\checkmark  & $\times$ &$\times$   \\ 
COIN~\cite{tang2019coin} & Diverse & 11,827 & 476  & \checkmark & $\times$ & $\times$ & $\times$ & $\times$ \\
YouCook2~\cite{zhou2018towards_youcook2} & Cooking & 2,000 & 176  & \checkmark & $\times$ & $\times$ & $\times$ & $\times$ \\
IKEA~\cite{ben2021ikea} & Assembly &1,113 &35.3 &\checkmark  &$\times$ &$\times$ &$\times$ &$\times$  \\
HoloAssist~\cite{wang2023holoassist} & Object Manip. & 2,221 & 166 & $\times$ & $\times$ & \checkmark & \checkmark & $\times$ \\
CaptainCook4D~\cite{peddi2024captaincook4d} & Cooking & 384 & 94.5 & $\times$ & $\times$ & \checkmark & $\times$ & $\times$ \\
QEVD-Fit-Coach~\cite{panchal2024say_QEVD} & Fitness  & 223 & 13.5  & \checkmark & \checkmark & \checkmark & \checkmark & $\times$ \\
Assembly101~\cite{sener2022assembly101} &  Assembly & 4,321 & 513 & $\times$ & $\times$ & \checkmark & \checkmark & $\times$ \\
EgoPER~\cite{egoper_cvpr24} & Cooking & 386  & 28 & \checkmark & $\times$ & \checkmark & $\times$ & $\times$ \\
\midrule
\textbf{Ours} & Diverse & 2,458  &  223.7
& \checkmark & \checkmark & \checkmark & \checkmark & \checkmark \\
\bottomrule
\end{tabular}}
\end{table*}

\section{Related Works}
\label{sec:related}

\subsection{Multimodal Large Language Models} 
Recent advances in Multimodal Large Language Models (MLLMs) have substantially broadened the scope of visual understanding~\cite{liu2023referring, liu2025language, liu2026gensplat}. 
Proprietary models such as GPT-4o~\cite{hurst2024gpt} and Gemini 3 Pro~\cite{gemini3pro2025} set new standards across diverse tasks spanning video understanding~\cite{fu2025video,yu2019activitynet}, document recognition~\cite{mathew2021docvqa,fu2024ocrbenchv2improvedbenchmark}, and mathematical reasoning~\cite{wang2024measuring,zhang2024mathverse}. Open-source counterparts, including InternVL 3.5~\cite{wang2025internvl3}, MiniCPM-V 4.5~\cite{yu2025minicpm}, Qwen3~\cite{xu2025qwen3omnitechnicalreport, bai2025qwen3vltechnicalreport, Qwen3-VL}, and DeepSeek-VL2~\cite{wu2024deepseek}, have rapidly narrowed the gap. However, these models are predominantly designed for offline analysis of pre-recorded content. Deploying them as interactive assistants that must observe, reason, and respond within a continuous video stream remains a largely open challenge.

\subsection{Streaming MLLMs and Benchmarks} 
Early video MLLMs digest a clip offline and respond only after the full video, ill-suited to assistance that must react as events unfold. One line of streaming models thus targets \emph{when} and \emph{how} to respond live: VideoLLM-online~\cite{chen2024videollm} interleaves perception and generation, Dispider~\cite{qian2025dispider} disentangles perception, decision, and reaction, and StreamMind~\cite{ding2025streammind} reaches full frame rate via event-gated cognition. Another scales to long, unbounded streams (StreamingVLM~\cite{xu2025streamingvlm}, LiveStar~\cite{yang2025livestar}) or retrofits offline backbones into streaming responders (StreamBridge~\cite{wang2025streambridge}, MMDuet2~\cite{wang2025mmduet2}). Orthogonally, memory-centric agents (M3-Agent~\cite{long2025seeing_m3agent}, HippoMM~\cite{lin2026hippomm}) target \emph{post-hoc} QA over recorded video rather than proactive intervention. 
For evaluation, benchmarks such as StreamingBench~\cite{lin2024streamingbench}, OVO-Bench~\cite{niu2025ovo}, PhoStream~\cite{lu2026phostream}, OmniMMI~\cite{wang2025omnimmi}, SVBench~\cite{yang2025svbench}, SpaceVista~\cite{sun2025spacevista}, and X-Stream~\cite{sun2026x}
assess general streaming understanding and proactive reasoning, while ProactiveVideoQA~\cite{wang2025proactivevideoqa} further targets the user experience of proactive interaction.
GuideMe is fundamentally harder: beyond understanding the stream, the model must \emph{proactively} decide when to intervene, what to advise, and whether the user erred, demanding far stronger temporal reasoning and situational awareness.

\subsection{Datasets for Procedural Activities}
Procedural video datasets differ mainly in how much of the coaching loop they supervise, and none spans it from instruction to recovery (Tab.~\ref{tab:dataset_comparison_new}).
The largest group pairs step-level instructions with expert demonstrations, ranging from cooking and assembly recordings~\cite{damen2020epic,zhou2018towards_youcook2,tang2019coin,ben2021ikea} to large-scale egocentric collections~\cite{miech2019howto100m,bao2023can_WTAG,grauman2022ego4d,grauman2024egoexo,huang2024egoexolearn,yang2025egolife}. Because every step is executed correctly, these data show what a good action looks like but never how to recover from a flawed one.
Error-aware datasets~\cite{sener2022assembly101,peddi2024captaincook4d,jang2019epic_tent,egoper_cvpr24} restore the missing failure signal, yet stop at detecting mistakes and leave the user without guidance on how to fix them.
Closest to genuine assistance, HoloAssist~\cite{wang2023holoassist} records real instructor corrections and QEVD~\cite{panchal2024say_QEVD} adds timed, action-specific feedback; even so, neither closes the loop, since later guidance is scripted in advance rather than conditioned on whether the user actually recovered, and QEVD stays within a single fitness domain.
All of these corpora, moreover, are curated offline. GuideMe is the first to combine four domains, model the full instruct-observe-correct cycle, and evaluate it under an online streaming protocol.

\section{The GuideMe Benchmark}
\label{sec:benchmark}

In this section, we present the GuideMe benchmark, covering task definition (Sec.~\ref{subsec:define}), annotation pipeline (Sec.~\ref{subsec:pipeline}), dataset statistics (Sec.~\ref{subsec:data_stat}), inference pipeline (Sec.~\ref{subsec:inference}), and evaluation protocol (Sec.~\ref{subsec:evaluation}).

\subsection{Task Definition}
\label{subsec:define}

We formulate interactive procedural assistance as an online streaming task under a strict causal setting. Given a continuous video stream of user execution, the assistant observes frames sequentially and can only access visual evidence up to the current timestamp, without any future information.
At each time step $t$, the assistant takes as input the recent video within a fixed sliding window and the full dialogue history accumulated so far. Based on this partial observation, it must decide whether to remain silent or to generate a guidance response.

If a response is required, the assistant produces one of three types of output:
(i)~a \emph{next-step instruction} directing the upcoming action,
(ii)~\emph{completion feedback} confirming that the current action has been successfully finished, or
(iii)~\emph{corrective feedback} when an execution error is detected, specifying both the mistake and how to fix it.
The user's subsequent action then updates the streaming context, forming a closed-loop interaction within a single session.

\subsection{Annotation Pipeline}
\label{subsec:pipeline}

\begin{figure*}[t]
\centering
\includegraphics[width=1\linewidth]{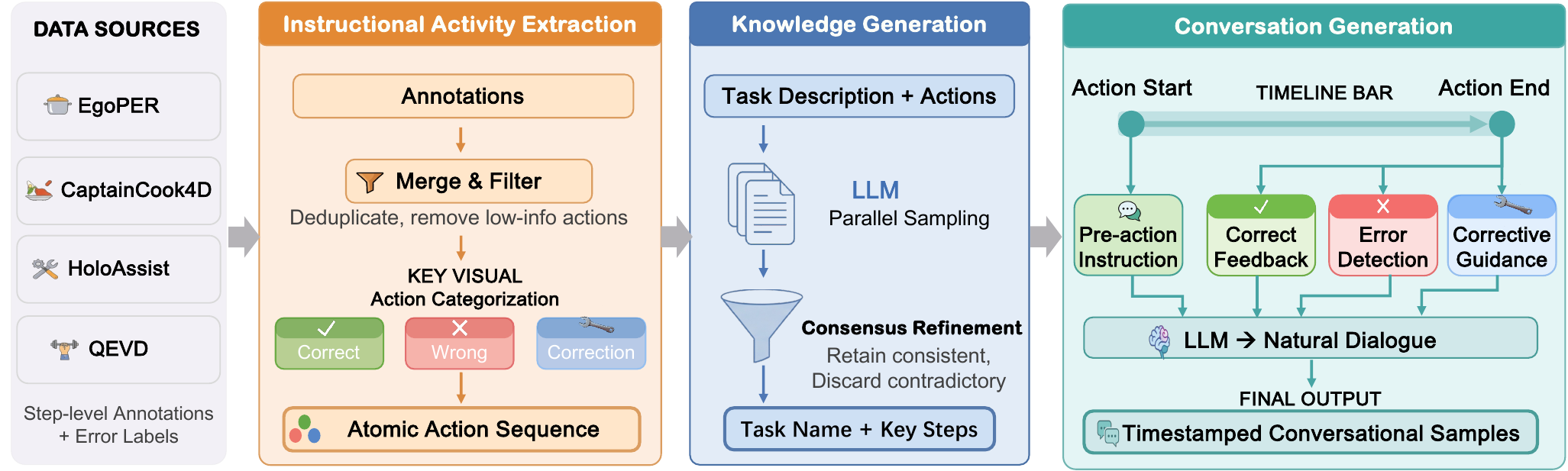}
\caption{\textbf{Annotation pipeline of GuideMe.} The process integrates three stages: instructional activity extraction with error categorization, knowledge generation from the extracted activities, and conversation generation. 
}
\label{fig:pipeline}
\end{figure*}

We construct GuideMe through a three-stage automated annotation pipeline, illustrated in Fig.~\ref{fig:pipeline}. Our data sources include EgoPER~\cite{egoper_cvpr24}, CaptainCook4D~\cite{peddi2024captaincook4d}, HoloAssist~\cite{wang2023holoassist}, and QEVD~\cite{panchal2024say_QEVD}. They span cooking, object manipulation, daily-life tasks, and fitness, and provide fine-grained step-level annotations of actions, mistakes, and corrective behaviors. Starting from these annotations, the pipeline proceeds from activity extraction to knowledge-guided dialogue generation.

\subsubsection{Instructional Activity Extraction.}
We first extract a complete and ordered set of task instructions from the raw step-level annotations, where each instruction corresponds to an independent \emph{atomic action} that can be executed, verified, or queried within a single dialogue turn. For datasets with fine-grained annotations, we merge duplicated or consecutive repeated actions and filter low-information ones (\eg, \textit{hold}, \textit{rotate}) to ensure each entry conveys meaningful task progress. Each resulting action is categorized as a \textbf{correct action} (following the intended procedure), a \textbf{wrong action} (deviating from the workflow), or a \textbf{correction action} (rectifying a prior mistake), according to the dataset annotations. \fang{These labels are inherited from each source's task graph or human annotations, which already encode parallel valid orderings.}
This categorization enables modeling of both ideal procedural execution and realistic error-and-recovery behaviors, which form the hallmark of our closed-loop evaluation.

\subsubsection{Knowledge Generation.}
Given the filtered instructional activity sequence from the extraction step and the original task description, we instruct the LLM to (1)~infer a descriptive task name and (2)~generate high-level procedural knowledge in the form of numbered key steps. Because the input actions may still include erroneous or corrective steps, the prompt emphasizes that the LLM must rely on commonsense reasoning to identify essential and logically consistent ones.

To improve robustness, we perform parallel sampling ($N{=}10$) and aggregate all candidates in a second-round refinement prompt, where steps consistently appearing across candidates are retained and contradictory ones discarded.

\subsubsection{Conversation Generation.}
In the final stage, we generate instructional dialogues aligned with the \emph{atomic action} boundaries, guided by the task description and generated procedural knowledge. For each action, the assistant produces utterances at two temporal points:
(1) at the \emph{start}, a {pre-action instruction} directs the upcoming step; 
(2) at the \emph{end}, the assistant issues either {correct execution feedback} confirming successful completion, or {error detection with corrective guidance} that identifies the mistake at the end timestamp with explicit corrective instructions for the proper procedure.

The source of correction instructions varies by dataset. HoloAssist provides explicit correction annotations, while CaptainCook4D and EgoPER require deriving corrections from their task graphs: when an action deviates from the expected node, we identify the discrepancy and use the correct subsequent step as the corrective instruction.

We serialize each video's atomic instructions in temporal order, with timestamps attached to each structured entry (\eg, ``\texttt{[05:32]} \textit{Tighten the screw on the left panel}''). Because long videos may contain hundreds of entries, we split the sequence into \emph{clips} of at most 200 entries, each conditioned on global context (task summary and high-level steps) and local context (fine-grained sub-steps in the current temporal range). The LLM then transforms these structured clips into natural conversational dialogues.

\subsection{Dataset Statistics and Analysis} \label{subsec:data_stat}

\begin{figure*}[t]
\centering
\includegraphics[width=1\linewidth]{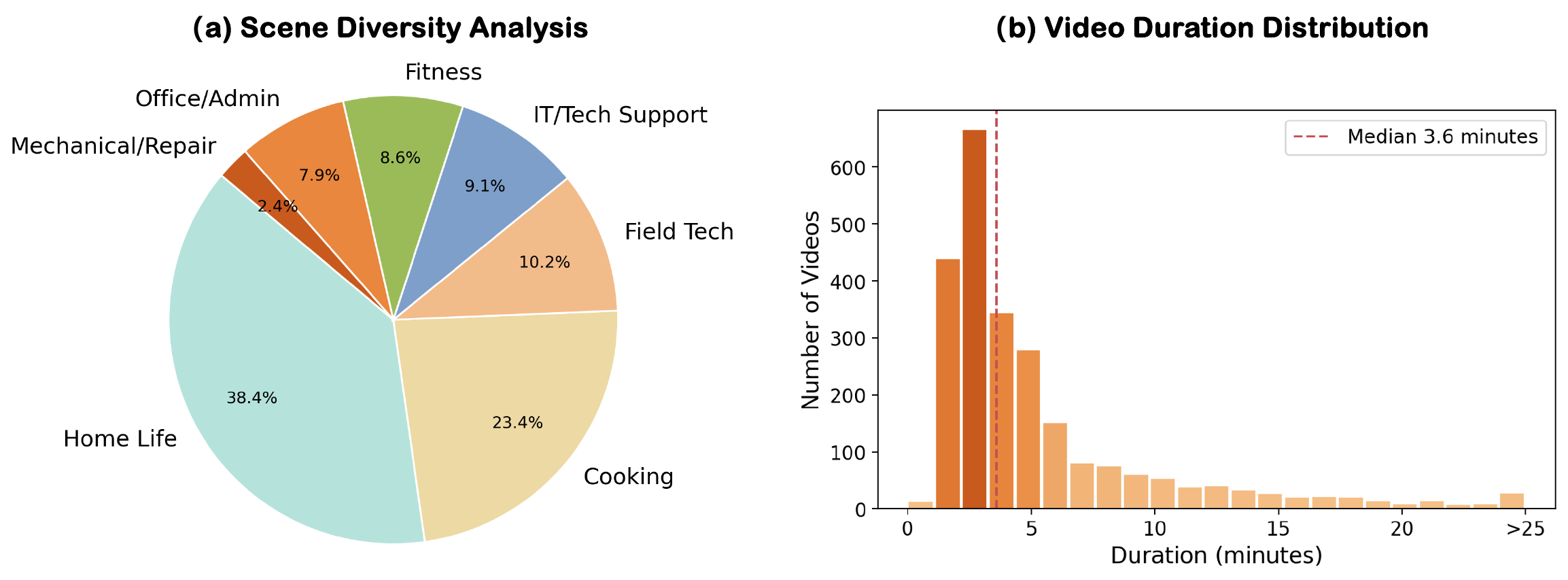}
\caption{\textbf{Dataset statistics of GuideMe.} (a)~Scene Diversity: Distribution across seven categories, dominated by Home Life (38.4\%) and Cooking (23.4\%), with technical domains (Field Tech and IT Support) comprising 38.2\%. 
(b)~Duration Distribution: Temporal spread of 2,458 videos with a median of 3.6 minutes. The distribution includes long-form sequences up to 41.2 minutes, providing challenging samples for evaluating long-range temporal reasoning.
}
\label{fig:data_statis}
\end{figure*}

The final dataset contains 2,458 video instances with a combined duration of 223.7 hours, drawn from CaptainCook4D\cite{peddi2024captaincook4d} (295 videos), EgoPER~\cite{egoper_cvpr24} (280), QEVD\cite{panchal2024say_QEVD} (212), and HoloAssist ~\cite{wang2023holoassist} (1,671). Video lengths range from 0.5 to 41.2 minutes (average 5.5\,min, median 3.6\,min), yielding 47,775 streaming interaction samples across cooking, daily tasks, fitness, and embodied assistance, of which 9,876 form the held-out test set used for all evaluations. \fang{GuideMe is split into \textbf{GuideMe-Train} (1,985 videos, 177.0 hours) and \textbf{GuideMe-Test} (473 videos, 46.7 hours); all reported evaluations use GuideMe-Test, while GuideMe-Train supports task-specific adaptation and is used only for the fine-tuned Qwen3-VL-8B result in Tab.~\ref{tab:main_results}.}

As shown in Fig.~\ref{fig:data_statis}, the domain distribution bridges simple daily routines and complex professional tasks: Home Life and Cooking form the foundational core, while Field Tech and IT Support provide the variety required for generalized evaluation. Temporally, the high density around the 3.6-minute median captures concise goal-oriented actions, while the extension to 41.2 minutes provides the depth needed for evaluating long-range reasoning and multi-step planning.

\subsection{Inference Pipeline}\label{subsec:inference}

We design an inference pipeline that mirrors the constraints of real-time interactive assistance. A single task-guidance query is issued once at the start of the video; thereafter the model consumes the stream strictly in chronological order and must \emph{proactively} decide when to intervene, with no external trigger. This departs fundamentally from question-answering benchmarks~\cite{lin2024streamingbench,niu2025ovo,wang2025proactivevideoqa}, which issue queries at fixed timestamps and only evaluate response timing; our pipeline instead tests whether a model can determine \emph{when} to speak at all.

Inference is run at discrete timestamps, for which we consider two sampling strategies. \textbf{Dense} sampling queries the model at a fixed 1\,s interval across the entire stream, so silent timestamps vastly outnumber the ground-truth interventions. \textbf{Anchor-Based} sampling (our default) instead queries at every ground-truth intervention plus an equal number of silent timestamps, yielding a balanced mix of speak and stay-silent cases. We compare the two in Sec.~\ref{sec:ablation}.

At each timestamp, the model observes the most recent 60 seconds of video through a sliding window~\cite{lu2026aura, lu2026phostream}, together with the full textual dialogue history (\ie, all previous assistant utterances and their timestamps). The sliding window bounds visual memory and enforces the online setting.
Given this context, the model must decide whether to remain silent or intervene. It outputs the single word \texttt{"Silent"} when no guidance is needed, and generates the appropriate instruction or feedback once sufficient evidence is available (\eg, when the user completes an action or commits an error).

\subsection{Evaluation Protocol}\label{subsec:evaluation}

Evaluating interactive streaming guidance requires assessing not only \emph{what} the model says but also \emph{when} it chooses to speak or remain silent. Traditional \fang{streaming metrics such as TimeDiff~\cite{chen2024videollm} presume a rigid one-to-one alignment to the reference, and LM-PPL~\cite{chen2024videollm} needs token probabilities that closed API models do not expose. Crucially, none of them evaluates the silence-versus-speak decision central to proactive guidance.} We address this with a three-component evaluation framework: (1)~temporal-semantic bipartite matching for sequence-level alignment, (2)~LLM-as-a-Judge scoring for content quality, and (3)~behavioral classification for instance-level intervention timing.

\subsubsection{Temporal-Semantic Bipartite Matching.}
We define the generated sequence as $\mathcal{G} = \{(g_i, t_i)\}_{i=1}^N$ and the reference sequence as $\mathcal{R} = \{(r_j, \hat{t}_j)\}_{j=1}^M$, where $g_i$ and $r_j$ denote the textual content of the $i$-th generated and $j$-th reference response, and $t_i$, $\hat{t}_j$ are their corresponding timestamps. 
To ensure strict temporal localization, we partition the video into $K$ segments $\{\mathcal{S}_k\}_{k=1}^K$ based on the midpoints of consecutive reference timestamps: $B_j = \frac{\hat{t}_j + \hat{t}_{j+1}}{2}$. This partitioning ensures that each segment $\mathcal{S}_k$ contains exactly one active (non-silent) reference event, providing a constrained search space for matching.

Within each segment $\mathcal{S}_k$, we seek the optimal assignment between $\mathcal{G}$ and $\mathcal{R}$ by solving a minimum weight bipartite matching problem, inspired by \cite{carion2020end, cheng2022masked}.
We define a cost matrix $\mathbf{C}$, where the cost $C_{i,j}$ for pairing $g_i$ with $r_j$ integrates textual alignment and temporal distance:
$C_{i,j} = \mathcal{L}_{\text{text}}(g_i, r_j) + \mathcal{L}_{\text{dist}}(t_i, \hat{t}_j)$.
The Textual Cost $\mathcal{L}_{\text{text}}$ is:
\begin{equation}
\mathcal{L}_{\text{text}}(g_i, r_j) =
\begin{cases}
0, & \text{if } g_i, r_j \in \text{Silence} \\
1 - |\cos(\mathbf{e}_i, \mathbf{\hat{e}}_j)|, & \text{otherwise} \\
\end{cases}
\end{equation}
where $\mathbf{e}_i, \mathbf{\hat{e}}_j$ are embeddings from a pre-trained Sentence-Transformer. The ``otherwise'' case captures all mismatches, including the model generating text during a silent ground-truth window or vice versa.
The temporal Distance Cost $\mathcal{L}_{\text{dist}}$ penalizes time-shifted responses via a Gaussian-decay function:
\begin{equation}
    \mathcal{L}_{\text{dist}}(t_i, \hat{t}_j) = 1 - \exp\left(-\sigma |t_i - \hat{t}_j|^2\right),
\end{equation}

Let $\mathcal{A}$ denote the optimal assignment obtained from the bipartite matching. To reward models that capture the essence of an action even if the wording differs from the reference, we define soft versions of precision and recall that use continuous semantic scores $S_{i,j}$ as weights instead of binary counts:
\begin{equation}
\text{sPrecision} = \frac{\sum_{(i,j) \in \mathcal{A}} S_{i,j}}{\tilde{N}_{\text{gen}}}, \quad 
\text{sRecall} = \frac{\sum_{(i,j) \in \mathcal{A}} S_{i,j}}{\tilde{M}_{\text{ref}}},
\end{equation}
where $S_{i,j} = |\cos(\mathbf{e}_i, \mathbf{\hat{e}}_j)|$ denotes the cosine similarity between embeddings $\mathbf{e}_i$ and $\mathbf{\hat{e}}_j$. The terms $\tilde{N}_{\text{gen}}$ and $\tilde{M}_{\text{ref}}$ represent the total count of active talk events in the generated sequence $\mathcal{G}$ and reference sequence $\mathcal{R}$, respectively. Their harmonic mean gives the Soft-$F_1$ score:
$
\text{sF}_1 = \frac{2 \cdot \text{sPrecision} \cdot \text{sRecall}}{\text{sPrecision} + \text{sRecall}},
$
which provides a unified, threshold-free measure that balances semantic accuracy with temporal coverage.

\subsubsection{LLM-as-a-Judge.}
After bipartite matching identifies the best-aligned prediction-reference pairs, we further evaluate the response quality of each matched pair using an LLM-as-a-Judge~\cite{zheng2023judging}. The judge is given the query context, the model prediction, and the reference answer, and assigns an integer score from 0 to 5 based on relevance, factual plausibility, and causal reasoning. We linearly rescale the score to 0--100 and average it over all matched pairs, yielding \textbf{Score$_m$}. Unlike sPrecision, sRecall, and sF$_1$, which measure temporal-semantic coverage, Score$_m$ isolates the quality of model responses at successfully matched intervention points.

\subsubsection{Behavioral Classification.}

While bipartite matching evaluates sequence-level alignment, we also need to assess whether the model intervenes at the right individual moments. For each video, we construct two complementary evaluation instances: (1)~\textbf{assistant-anchored instances}, where each ground-truth intervention timestamp serves as a temporal anchor and the model must produce the appropriate guidance; and (2)~\textbf{silent instances}, sampled from intervals between consecutive interventions (at least 5\,s from any anchor), where the model should remain silent. The two types are balanced in number.

Each prediction is categorized into one of four mutually exclusive outcomes: 
Correct Silent (\textbf{CS}, $\uparrow$), where the model correctly remains silent for a silent instance; 
False Alarm (\textbf{FA}, $\downarrow$), where the model speaks during a silent instance; 
No Response (\textbf{NR}, $\downarrow$), where the model fails to respond at an intervention-anchored timestamp; 
and 
Partly Correct (\textbf{PC}, $\uparrow$), where the model produces a response at an intervention-anchored timestamp. 
These categories capture two complementary timing failures: \emph{over-intervention}, reflected by FA, and \emph{under-intervention}, reflected by NR.

We further summarize these behavior outcomes with an aggregate response-behavior metric, the \textbf{Score}. CS is assigned a score of 100, as the model correctly remains silent; FA and NR are assigned a score of 0, corresponding to false intervention and erroneous silence, respectively; and each PC instance receives an LLM-as-a-Judge~\cite{zheng2023judging} quality score, linearly scaled to the range of 0--100. The \textbf{Score} reported in our tables is the average of these values over all evaluation instances, rewarding both correct silence and high-quality interventions while penalizing false alarms and missed responses.

\section{Experiments}
\label{sec:experiments}

In this section, we present experiments on GuideMe. We describe the experimental setup (Sec.~\ref{sec:exp_set}), report baseline and visual results (Sec.~\ref{sec:res_main} and Sec.~\ref{sec:res_qualitative}), and conduct ablation studies on inference settings (Sec.~\ref{sec:ablation}). Additional per-domain breakdowns and prompt details are provided in the supplementary material.

\subsection{Experiment Setup}\label{sec:exp_set}

We evaluate three categories of zero-shot baselines on GuideMe using the Inference Pipeline (Sec.~\ref{subsec:inference}): (1)~proprietary MLLMs (Doubao-Seed-1.8~\cite{seed2026vision}, GPT-5.2~\cite{singh2025openai}, Gemini~3~Pro~\cite{gemini3pro2025}, Gemini~3.1~Pro~\cite{gemini3.1pro2025}), (2)~open-source MLLMs (Qwen2.5-7B~\cite{Qwen2.5-VL}, Qwen3-VL-8B~\cite{Qwen3-VL}, Qwen3-VL-30B-A3B~\cite{Qwen3-VL}, Qwen3.5-397B-A17B~\cite{qwen35blog}), and (3)~open-source streaming MLLMs (VideoLLM-online~\cite{chen2024videollm}, Dispider~\cite{qian2025dispider}, LiveStar~\cite{yang2025livestar}, MMDuet2~\cite{wang2025mmduet2}). We additionally report Qwen3-VL-8B fine-tuned on the GuideMe training split to assess whether task-specific adaptation improves streaming guidance.
\fang{By default we provide the ground-truth dialogue history at each step so that scores isolate each model's single-step timing and content decisions from error accumulation across turns.}
We report the two metric families defined in Sec.~\ref{subsec:evaluation}: \textit{Temporal Alignment} (sPrecision, sRecall, sF$_1$, Score$_m$) and \textit{Response Behavior} (CS, FA, NR, PC, Score).

\begin{table*}[t]
    \centering
    \small
    \setlength{\tabcolsep}{2pt}
    \caption{\textbf{Main evaluation results on GuideMe.} We assess model responses from two complementary perspectives: \textit{Temporal Alignment} evaluates whether the model provides guidance at the appropriate time with relevant content, including Score$_m$ for the LLM-as-a-Judge quality of matched prediction-reference pairs. \textit{Response Behavior} categorizes each instance into Correct Silent (\textbf{CS}, $\uparrow$), No Response (\textbf{NR}, $\downarrow$), False Alarm (\textbf{FA}, $\downarrow$), and Partly Correct (\textbf{PC}, $\uparrow$), and summarizes them with \textbf{Score}. The best result in each category is \textbf{bolded}. $^\dagger$ denotes fine-tuning on the GuideMe training split.}
    \label{tab:main_results}
    \resizebox{0.98\textwidth}{!}{
    \begin{tabular}{l|c|ccc|c|cccc|c}
    \toprule
    \multirow{2}{*}{Model} & \multirow{2}{*}{Param.} 
    & \multicolumn{4}{c|}{Temporal Alignment} 
    & \multicolumn{5}{c}{Response Behavior} 
    \\
    \cmidrule(lr){3-6}\cmidrule(lr){7-11}
    & & sPrecision $\uparrow$ & sRecall $\uparrow$ & sF1 $\uparrow$ 
    & Score$_m$ $\uparrow$
    & CS $\uparrow$ & NR $\downarrow$ & FA $\downarrow$ & PC $\uparrow$ & Score $\uparrow$ \\
    \midrule
    \multicolumn{11}{l}{\textit{Proprietary Multimodal Models}} \\
    \midrule
    Doubao-Seed-1.8~\cite{seed2026vision} & - &30.6  & \textbf{47.5}  &36.8 &63.1  & 8.8  & 6.1  & 34.8 &44.2 &38.8 \\
    GPT-5.2~\cite{singh2025openai} &- &30.2 &36.8  &32.5 &64.4  &13.4 &16.8 &30.6 &39.2 &38.9 \\
    Gemini 3 Pro~\cite{gemini3pro2025} & - & 39.3 &36.9 &36.5 &\textbf{66.6}  & 17.5 & 22.0 &19.0  & 41.5 &44.8 \\
    Gemini 3.1 Pro~\cite{gemini3.1pro2025} & - & 29.3 &46.9 &35.7 &61.5   & 7.4 & 6.3 & 36.4 &49.9  & 37.8 \\
    \midrule
    \multicolumn{11}{l}{\textit{Open-source Multimodal Models}} \\
    \midrule
    Qwen2.5-7B~\cite{Qwen2.5-VL} & 7B  &30.1 &30.9 &29.3 &57.5 & 19.7 &20.4  &25.4  &34.5  &37.5 \\
    Qwen3-VL-8B~\cite{Qwen3-VL} & 8B & 29.7 &46.8 &35.7 &62.0 &5.2  &6.7  &37.7  &50.5  &31.9 \\
    Qwen3-VL-30B-A3B~\cite{Qwen3-VL} & 30B &29.5 &46.9 &35.6 &61.6 &6.7  &6.0  &36.5  &50.9  &33.3 \\
    Qwen3.5-397B-A17B~\cite{qwen35blog} & 397B &38.4 &20.2 &21.7 &64.6 &35.2 &37.9 &7.9  &19.0  &\textbf{45.7} \\
    \midrule
    \multicolumn{11}{l}{\textit{Open-source Multimodal Streaming Models}} \\
    \midrule
    VideoLLM-online~\cite{chen2024videollm} & 8B & 22.3 & 41.1 & 28.7 &40.8 & 0.0 & \textbf{0.0} & 43.6 & \textbf{56.4} & 22.6 \\
    Dispider~\cite{qian2025dispider} & 7B & 1.0 & 0.1 & 0.1 &46.7 & \textbf{43.6} & 56.3 & \textbf{0.0} & 0.1 & 43.6 \\
    LiveStar~\cite{yang2025livestar} & 8B & 20.7 & 19.0 & 19.1 &14.4 & 21.2 & 24.4 & 22.2 & 32.3 & 25.0 \\
    MMDuet2~\cite{wang2025mmduet2} & 3B & 3.3 & 0.2 & 0.4 &43.3 & 41.7 & 57.9 & 0.2 & 0.2 & 41.8 \\
    \midrule
    \multicolumn{11}{l}{\textit{Fine-tuned on GuideMe Train Set}} \\
    \midrule
    Qwen3-VL-8B$^\dagger$~\cite{Qwen3-VL} & 8B & \textbf{40.3} & 43.7 & \textbf{39.7} &58.1 & 19.5 & 20.6 & 24.5 & 35.3 & 42.1 \\
    \bottomrule
    \end{tabular}}
    \end{table*}

\subsection{Quantitative Results}\label{sec:res_main}

Tab.~\ref{tab:main_results} reports aggregate results on the full GuideMe test set, while Tab.~\ref{tab:response_type_main} further stratifies representative models by ground-truth response category to analyze where failures occur. Overall, no zero-shot baseline provides reliable streaming coaching: models either intervene too often, remain silent too often, or fail to detect when an intervention should change from routine guidance to error correction.

\noindent\textbf{Response frequency trades off with quality.}
The results show a clear split between aggressive models, which respond frequently but raise many false alarms, and conservative models, which avoid false alarms but miss required interventions. For example, Qwen3-VL-30B-A3B reaches 50.9\% PC but only 33.3 Score, while Qwen3.5 obtains the highest Score (45.7) largely by speaking less. Score$_m$ further shows that matched-response quality is relatively close among non-streaming MLLMs (57.5--66.6), suggesting that deciding \emph{when} to intervene is the more limiting factor.

\noindent\textbf{Scaling and streaming pretraining are insufficient.}
They further show that general scaling and streaming-oriented pretraining do not solve the task. Qwen3-VL-8B and Qwen3-VL-30B-A3B obtain nearly identical sF1 scores (35.7 vs.\ 35.6), while the much larger Qwen3.5-397B-A17B performs worse on temporal alignment (sF1 21.7). Streaming-trained models show the same issue more severely: VideoLLM-online over-responds (FA 43.6, nearly never silent), whereas Dispider and MMDuet2 collapse to near-silence (sF1 $<$ 1, NR $>$ 55\%). This indicates that knowing \emph{when} to speak, not only how to describe video content, is the core missing capability.

\noindent\textbf{Fine-tuning improves alignment but shifts behavior.}
Fine-tuning on GuideMe improves Qwen3-VL-8B at the sequence level, raising sPrecision from 29.7 to 40.3 and sF1 from 35.7 to 39.7. At the instance level, however, the model becomes more conservative: CS increases from 5.2 to 19.5 and FA decreases from 37.7 to 24.5, indicating better silence calibration, but PC also drops from 50.5 to 35.3 and NR rises from 6.7 to 20.6. At the content level, Score improves from 31.9 to 42.1, yet remains below the strongest zero-shot models. Thus, the training split supports task-specific adaptation, but robust closed-loop coaching still requires jointly improving temporal alignment, intervention decisions, and response quality.

\noindent\textbf{Error correction remains the hardest behavior.}
Tab.~\ref{tab:response_type_main} shows that both Gemini~3~Pro and Doubao-Seed-1.8 perform best on Instruction and degrade on Error and Correction. For Gemini~3~Pro, sF1 drops from 42.0 on Instruction to 33.8 on Error and 34.3 on Correction, while Score drops from 33.0 to 20.2 and 23.1. Doubao follows the same trend, with sF1 dropping from 49.6 to 41.9 and 42.7. Current MLLMs are therefore stronger at routine step guidance than at timely error detection and corrective intervention.

\begin{table}[t]
    \centering
    \scriptsize
    \setlength{\tabcolsep}{3pt}
    \caption{\textbf{Per-category breakdown in the main results.} Metrics are stratified by ground-truth response category on the full test set.}
    \label{tab:response_type_main}
    \resizebox{\linewidth}{!}{
    \begin{tabular}{l|cccc|cccc|cccc}
    \toprule
    \multirow{2}{*}{Model} & \multicolumn{4}{c|}{Instruction} & \multicolumn{4}{c|}{Error} & \multicolumn{4}{c}{Correction} \\
    \cmidrule(lr){2-5}\cmidrule(lr){6-9}\cmidrule(lr){10-13}
    & sF$_1\uparrow$ & NR$\downarrow$ & PC$\uparrow$ & Score$\uparrow$ & sF$_1\uparrow$ & NR$\downarrow$ & PC$\uparrow$ & Score$\uparrow$ & sF$_1\uparrow$ & NR$\downarrow$ & PC$\uparrow$ & Score$\uparrow$ \\
    \midrule
    Gemini 3 Pro~\cite{gemini3pro2025} & 42.0 & 33.3 & 66.7 & 33.0 & 33.8 & 39.6 & 60.4 & 20.2 & 34.3 & 41.7 & 58.3 & 23.1 \\
    Doubao-Seed-1.8~\cite{seed2026vision} & 49.6 & 8.9 & 91.1 & 41.9 & 41.9 & 14.8 & 85.2 & 26.2 & 42.7 & 15.0 & 85.0 & 30.0 \\
    \bottomrule
    \end{tabular}
    }
\end{table}

\definecolor{clrGT}{RGB}{210,120,20}
\definecolor{clrUser}{RGB}{70,70,70}
\begin{figure*}[!t]
  \centering
  \resizebox{0.98\linewidth}{!}{%
  \begin{minipage}{\linewidth}
  \centering

  \includegraphics[width=1\linewidth]{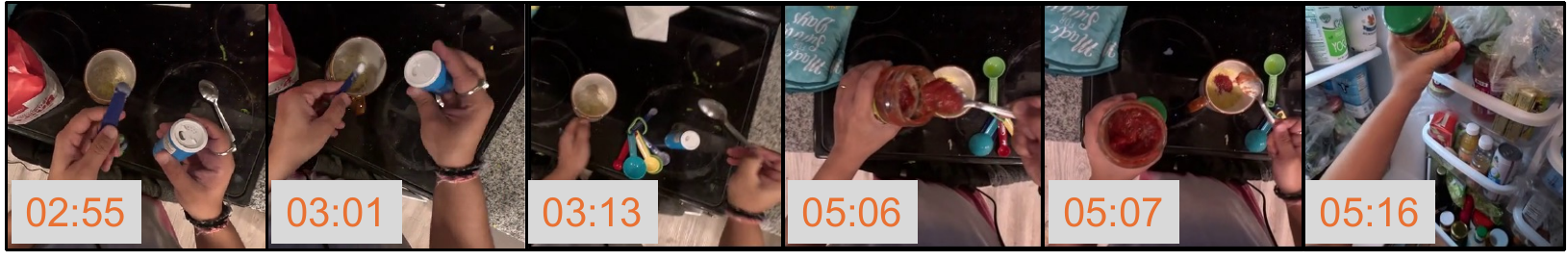}

  \vspace{1pt}

  \begin{tcolorbox}[
    colback=clrUser!4,
    colframe=clrUser!40,
    boxrule=1pt,
    arc=3pt,
    boxsep=1pt,
    left=4pt, right=4pt, top=4pt, bottom=4pt,
    fonttitle=\scriptsize\bfseries,
    title={\faUser~User}
  ]
  {\scriptsize
  I'm going to make a microwave mug pizza. Can you walk me through the steps and let me know if I make any mistakes along the way?
  }
  \end{tcolorbox}

  \vspace{-16pt}
  \begin{center}
  \textcolor{clrGT}{\tiny$\bullet$\;\;$\bullet$\;\;$\bullet$}
  \end{center}
  \vspace{-16pt}

  \begin{tcolorbox}[
    colback=clrGT!4,
    colframe=clrGT!40,
    boxrule=1pt,
    arc=3pt,
    left=4pt, right=4pt, top=3pt, bottom=3pt,
    fonttitle=\scriptsize\bfseries,
    title={Assistant (03:01)}
  ]
  {\scriptsize
  \textcolor{clrGT}{\textbf{GT:}} \textcolor{clrGT}{Hold on a second. It looks like you're about to add the salt, but please make sure you're using a clean, dry measuring spoon so the ingredients don't stick.}

  \vspace{4pt}
  \textbf{Doubao-Seed-1.8:} Great, herbs are added. Now go ahead and add a tablespoon of mozzarella cheese to the mug.

  \vspace{4pt}
  \textbf{GPT-5.2:} \textit{[Silent]}

  \vspace{4pt}
  \textbf{Gemini 3.1 Pro:} Perfect. Now add 1/16 teaspoon of baking soda.

  \vspace{4pt}
  \textbf{Qwen2.5-7B:} Now, add 2 tablespoons of pizza sauce.

  \vspace{4pt}
  \textbf{MMDuet2:} \textit{[Silent]}

  \vspace{4pt}
  \textbf{Qwen3-VL-8B (SFT):} That's the right amount of herbs. Now, add 1/16 teaspoon of black pepper to the mix.
  }
  \end{tcolorbox}

  \vspace{-16pt}
  \begin{center}
  \textcolor{clrGT}{\tiny$\bullet$\;\;$\bullet$\;\;$\bullet$}
  \end{center}
  \vspace{-16pt}

  \begin{tcolorbox}[
    colback=clrGT!4,
    colframe=clrGT!45,
    boxrule=1pt,
    arc=3pt,
    left=4pt, right=4pt, top=3pt, bottom=3pt,
    fonttitle=\scriptsize\bfseries,
    title={Assistant (05:07)}
  ]
  {\scriptsize
  \textcolor{clrGT}{\textbf{GT:}} \textcolor{clrGT}{Just a heads-up: it looks like you're reaching for the sauce, but make sure you use marinara, not salsa.}

  \vspace{4pt}
  \textbf{Doubao-Seed-1.8:} Great, you've grabbed the pizza sauce. Now add 2 tablespoons of it to the mug.

  \vspace{4pt}
  \textbf{GPT-5.2:} Now add 1 tablespoon of pizza sauce on top, do not stir it in. Spread it lightly.

  \vspace{4pt}
  \textbf{Gemini 3.1 Pro:} Now, spoon 1 tablespoon of marinara sauce onto the surface of the dough and spread it around.

  \vspace{4pt}
  \textbf{Qwen2.5-7B:} Microwave for 1 minute and 20 seconds.

  \vspace{4pt}
  \textbf{MMDuet2:} \textit{[Silent]}

  \vspace{4pt}
  \textbf{Qwen3-VL-8B (SFT):} It looks like you're trying to add the tomato sauce, but you're spilling some of it while you're at it.
  }
  \end{tcolorbox}

  \vspace{-16pt}
  \begin{center}
  \textcolor{clrGT}{\tiny$\bullet$\;\;$\bullet$\;\;$\bullet$}
  \end{center}
  \vspace{-12pt}

  \end{minipage}}

  \caption{\textbf{Qualitative example.} A microwave mug pizza task with two error-correction interventions. The ground-truth (GT) response is shown in {\color[RGB]{210,120,20}{orange}} alongside predictions from six representative models.}
  \label{fig:qualitative_main}
  \end{figure*}

\subsection{Qualitative Results}\label{sec:res_qualitative}

Fig.~\ref{fig:qualitative_main} shows a representative cooking example (a microwave mug pizza), comparing the ground-truth guidance with predictions from six models at two error-correction intervention points. The example highlights that models often continue routine instruction even when corrective feedback is needed. At 03:01, most responding models advance to later recipe steps instead of warning about the measuring spoon, while GPT-5.2 and MMDuet2 remain silent. At 05:07, Gemini~3.1~Pro correctly identifies marinara sauce, but several models either give generic next-step guidance or miss the correction.

\subsection{Ablation Studies}
\label{sec:ablation}

We conduct four ablations in two groups. Tab.~\ref{tab:ablation_combined} studies three inference-protocol settings on a randomly sampled 100-video subset (25 videos from each source dataset). The \textit{Dense} and \textit{Anchor} settings follow Sec.~\ref{subsec:inference}, and both use ground-truth (GT) dialogue history; Dense queries every 1\,s, while Anchor queries intervention timestamps and balanced silent timestamps. To further evaluate fully autonomous context, the \textit{w/o GT} setting keeps Anchor-Based sampling but replaces GT dialogue history with the model's own predictions. Tab.~\ref{tab:hyperparam} then analyzes Gemini~3~Pro with two sensitivity tests: under Anchor-Based sampling, we vary the sliding window size to control the visible visual context; under Dense sampling, we vary the sampling interval to control query frequency.

\noindent\textbf{Anchor-Based vs.\ Dense Sampling.}
Dense sampling exposes a strong silence imbalance: Doubao and Qwen2.5 collapse to silence (CS$+$NR $>$93\%), inflating Score through the silent bonus (\eg, Doubao 38.6$\to$90.3) while their coaching ability (PC) collapses. GPT-5.2 and Gemini~3~Pro instead over-respond at nearly half of all timestamps (FA $>$46\%). Anchor-Based sampling avoids this distributional bias, so we adopt it as the default.

\noindent\textbf{Ground-Truth vs.\ Predicted Context.}
Removing GT context consistently degrades every model: sF1 and Score drop while missed interventions (NR) rise sharply (\eg, Doubao sF1 $36.9\to20.0$, NR $6.1\to50.9$; Qwen2.5 Score $37.3\to26.3$), as early mistakes compound into an increasingly unreliable dialogue history. This confirms GT history as the fairer evaluation protocol and quantifies the headroom that remains for fully autonomous deployment.

\begin{table*}[t]
    \centering
    \small
    \setlength{\tabcolsep}{4pt}
    \caption{\textbf{Effect of sampling and dialogue-history settings.} The default setting ($^*$) uses Anchor-Based sampling with ground-truth (GT) dialogue history. We compare Dense and Anchor-Based sampling, both defined in Sec.~\ref{subsec:inference}, under GT dialogue history, and further evaluate \textbf{w/o GT} by replacing GT history with the model's own predictions. Results are evaluated on a randomly sampled 100-video subset (25 videos from each source dataset). The best result per model in each column is \textbf{bolded}.
    }
    \label{tab:ablation_combined}
    \resizebox{0.98\textwidth}{!}{
    \begin{tabular}{l|c |ccc |cccc|c}
    \toprule
    \multirow{2}{*}{Model} & \multirow{2}{*}{Setting}
    & \multicolumn{3}{c|}{Temporal Alignment}
    & \multicolumn{5}{c}{Response Behavior}
    \\
    \cmidrule(lr){3-5}\cmidrule(lr){6-10}
    & & sPrecision $\uparrow$ & sRecall $\uparrow$ & sF1 $\uparrow$
    & CS $\uparrow$ & NR $\downarrow$ & FA $\downarrow$ & PC $\uparrow$ & Score $\uparrow$ \\
    \midrule
    \multirow{3}{*}{Doubao-Seed-1.8~\cite{seed2026vision}}
        & Dense       & 23.1 & 26.3 & 21.9 & \textbf{89.3} & \textbf{4.4} & \textbf{4.7} & 1.6 & \textbf{90.3} \\
        & Anchor$^*$  & 30.6 & \textbf{47.9} & \textbf{36.9} & 7.8 & 6.1 & 34.3 & \textbf{51.9} & 38.6 \\
        & w/o GT      & \textbf{43.5} & 14.2 & 20.0 & 33.9 & 50.9 & 8.4 & 6.8 & 38.3 \\
    \midrule
    \multirow{3}{*}{GPT-5.2~\cite{singh2025openai}}
        & Dense       & 10.7 & \textbf{40.6} & 14.8 & \textbf{47.4} & \textbf{3.1} & 46.4 & 3.1 & \textbf{49.0} \\
        & Anchor$^*$  & 30.4 & 37.0 & \textbf{32.8} & 12.7 & 17.1 & 29.8 & \textbf{40.4} & 38.8 \\
        & w/o GT      & \textbf{31.7} & 17.9 & 21.6 & 26.1 & 43.3 & \textbf{16.1} & 14.4 & 34.1 \\
    \midrule
    \multirow{3}{*}{Gemini 3 Pro~\cite{gemini3pro2025}}
        & Dense       & 10.0 & \textbf{53.2} & 14.5 & \textbf{45.4} & \textbf{2.1} & 48.4 & 4.1 & \textbf{47.9} \\
        & Anchor$^*$  & \textbf{39.7} & 35.6 & \textbf{35.7} & 18.1 & 23.1 & 19.0 & \textbf{39.9} & 43.7 \\
        & w/o GT      & 35.7 & 25.1 & 28.0 & 27.6 & 36.4 & \textbf{14.6} & 21.3 & 39.1 \\
    \midrule
    \multirow{3}{*}{Qwen2.5-7B~\cite{Qwen2.5-VL}}
        & Dense       & 29.1 & 15.4 & 18.6 & \textbf{90.5} & 5.5 & \textbf{3.3} & 0.7 & \textbf{91.0} \\
        & Anchor$^*$  & \textbf{32.7} & \textbf{33.0} & \textbf{31.4} & 19.2 & \textbf{19.2} & 25.5 & \textbf{36.2} & 37.3 \\
        & w/o GT      & 16.3 & 25.5 & 17.1 & 13.8 & 27.6 & 23.4 & 35.1 & 26.3 \\
    \bottomrule
    \end{tabular}}
\end{table*}

\noindent\textbf{Sliding Window Size.}
Tab.~\ref{tab:hyperparam} shows that the 60\,s visual window provides a reasonable balance between fidelity and cost. Shrinking it to 30\,s lowers Score by 4.5 as procedural context is lost, while enlarging it to 120\,s yields only a marginal $+2.8$ gain at $2{\times}$ token cost.

\noindent\textbf{Sampling Interval.}
A 2\,s interval drops Score by 7.1 as it misses fast actions, whereas 0.5\,s gives comparable Score at $2{\times}$ cost, confirming 1\,s as the default. Overall, GuideMe's conclusions remain robust to these choices rather than being artifacts of a particular sampling configuration.

\begin{table*}[t]
\centering
\small
\setlength{\tabcolsep}{8pt}
\caption{\textbf{Sensitivity analysis with Gemini~3~Pro.} \textit{Sliding Window Size} under the Anchor-Based sampling strategy (default: 60\,s window) and \textit{Sampling Interval} under the Dense sampling strategy (default: 1\,s interval). The best result within each block is \textbf{bolded}.}
\label{tab:hyperparam}
\resizebox{0.882\textwidth}{!}{
\begin{tabular}{c|ccc|cccc|c}
\toprule
\multirow{2}{*}{Setting}
& \multicolumn{3}{c|}{Temporal Alignment}
& \multicolumn{5}{c}{Response Behavior}
\\
\cmidrule(lr){2-4}\cmidrule(lr){5-9}
& sPrecision $\uparrow$ & sRecall $\uparrow$ & sF1 $\uparrow$ & CS $\uparrow$ & NR $\downarrow$ & FA $\downarrow$ & PC $\uparrow$ & Score $\uparrow$ \\
\midrule
\multicolumn{9}{c}{\textit{Sliding Window Size under Anchor-Based Sampling}} \\
\midrule
30\,s          & 31.6 & 40.1 & 34.8 & 13.7 & 22.8 & 23.1 & 40.2 & 40.3 \\
60\,s          & 33.8 & \textbf{40.3} & 36.5 & 17.5 & 22.0 & 19.0 & \textbf{41.5} & 44.8 \\
120\,s         & \textbf{35.1} & 39.8 & \textbf{37.1} & \textbf{19.9} & \textbf{21.9} & \textbf{16.7} & \textbf{41.5} & \textbf{47.6} \\
\midrule
\multicolumn{9}{c}{\textit{Sampling Interval under Dense Sampling}} \\
\midrule
0.5\,s         & 3.6  & \textbf{56.5} & 5.2  & 44.9 & \textbf{1.0}  & 51.9 & 2.0  & 46.2 \\
1\,s           & 10.0 & 53.2 & 14.5 & \textbf{45.4} & 2.1  & \textbf{48.4} & 4.1  & \textbf{47.9} \\
2\,s           & \textbf{11.6} & 54.0 & \textbf{16.8} & 35.2 & 2.6  & 52.8 & \textbf{9.3}  & 40.8 \\
\bottomrule
\end{tabular}}
\end{table*}

\section{Conclusion}

This paper presents GuideMe, the first multi-domain streaming benchmark for evaluating closed-loop interactive task guidance. GuideMe comprises 2,458 videos spanning 223.7 hours across four domains with 47,775 interaction samples, built through a scalable three-stage annotation pipeline and assessed with a three-component evaluation framework: Temporal-Semantic Bipartite Matching for sequence-level alignment, Behavioral Classification for instance-level intervention timing, and LLM-as-a-Judge for content quality.
Across proprietary, open-source, and streaming models, we find that current MLLMs polarize into two failure modes: aggressive models intervene frequently but raise many false alarms, whereas conservative ones stay silent and miss most interventions. As a result, no model coaches reliably: the best-balanced Gemini~3~Pro still misses 22\% of needed interventions, while Qwen3.5 obtains the highest Score (45.7) despite correctly responding at only 19\% of intervention-anchored instances. This exposes a fundamental gap in the instruct-observe-correct cycle: models can \emph{describe} procedures, but they cannot yet \emph{coach} them. We hope GuideMe serves as a practical testbed for developing AI assistants that move beyond passive narration toward genuinely interactive procedural coaching.

\bibliographystyle{splncs04}
\bibliography{main}

@String(CVPR  = {IEEE Conf. Comput. Vis. Pattern Recog.})

@String(ICCV  = {Int. Conf. Comput. Vis.})

@String(ECCV  = {Eur. Conf. Comput. Vis.})

@String(NeurIPS = {Adv. Neural Inform. Process. Syst.})

@String(ICML  = {Int. Conf. Mach. Learn.})

@String(ICLR  = {Int. Conf. Learn. Represent.})

@String(AAAI  = {AAAI})

@String(CVPR  = {CVPR})

@String(ICCV  = {ICCV})

@String(ECCV  = {ECCV})

@String(NeurIPS = {NeurIPS})

@String(ICML  = {ICML})

@String(ICLR  = {ICLR})

@article{Qwen3-VL,
      title={Qwen3-VL Technical Report}, 
      author={Shuai Bai and Yuxuan Cai and Ruizhe Chen and Keqin Chen and Xionghui Chen and Zesen Cheng and Lianghao Deng and Wei Ding and Chang Gao and Chunjiang Ge and Wenbin Ge and Zhifang Guo and Qidong Huang and Jie Huang and Fei Huang and Binyuan Hui and Shutong Jiang and Zhaohai Li and Mingsheng Li and Mei Li and Kaixin Li and Zicheng Lin and Junyang Lin and Xuejing Liu and Jiawei Liu and Chenglong Liu and Yang Liu and Dayiheng Liu and Shixuan Liu and Dunjie Lu and Ruilin Luo and Chenxu Lv and Rui Men and Lingchen Meng and Xuancheng Ren and Xingzhang Ren and Sibo Song and Yuchong Sun and Jun Tang and Jianhong Tu and Jianqiang Wan and Peng Wang and Pengfei Wang and Qiuyue Wang and Yuxuan Wang and Tianbao Xie and Yiheng Xu and Haiyang Xu and Jin Xu and Zhibo Yang and Mingkun Yang and Jianxin Yang and An Yang and Bowen Yu and Fei Zhang and Hang Zhang and Xi Zhang and Bo Zheng and Humen Zhong and Jingren Zhou and Fan Zhou and Jing Zhou and Yuanzhi Zhu and Ke Zhu},
	  journal={arXiv preprint arXiv:2511.21631},
      year={2025}
}

@article{Qwen2.5-VL,
  title={Qwen2.5-VL Technical Report},
  author={Bai, Shuai and Chen, Keqin and Liu, Xuejing and Wang, Jialin and Ge, Wenbin and Song, Sibo and Dang, Kai and Wang, Peng and Wang, Shijie and Tang, Jun and Zhong, Humen and Zhu, Yuanzhi and Yang, Mingkun and Li, Zhaohai and Wan, Jianqiang and Wang, Pengfei and Ding, Wei and Fu, Zheren and Xu, Yiheng and Ye, Jiabo and Zhang, Xi and Xie, Tianbao and Cheng, Zesen and Zhang, Hang and Yang, Zhibo and Xu, Haiyang and Lin, Junyang},
  journal={arXiv preprint arXiv:2502.13923},
  year={2025}
}

@inproceedings{wang2023holoassist,
  title={Holoassist: an egocentric human interaction dataset for interactive ai assistants in the real world},
  author={Wang, Xin and Kwon, Taein and Rad, Mahdi and Pan, Bowen and Chakraborty, Ishani and Andrist, Sean and Bohus, Dan and Feniello, Ashley and Tekin, Bugra and Frujeri, Felipe Vieira and others},
  booktitle={ICCV},
  pages={20270--20281},
  year={2023}
}

@article{damen2020epic,
  title={The epic-kitchens dataset: Collection, challenges and baselines},
  author={Damen, Dima and Doughty, Hazel and Farinella, Giovanni Maria and Fidler, Sanja and Furnari, Antonino and Kazakos, Evangelos and Moltisanti, Davide and Munro, Jonathan and Perrett, Toby and Price, Will and others},
  journal={IEEE TPAMI},
  volume={43},
  number={11},
  pages={4125--4141},
  year={2020},
  publisher={IEEE}
}

@inproceedings{grauman2022ego4d,
  title={Ego4d: Around the world in 3,000 hours of egocentric video},
  author={Grauman, Kristen and Westbury, Andrew and Byrne, Eugene and Chavis, Zachary and Furnari, Antonino and Girdhar, Rohit and Hamburger, Jackson and Jiang, Hao and Liu, Miao and Liu, Xingyu and others},
  booktitle={CVPR},
  pages={18995--19012},
  year={2022}
}

@inproceedings{grauman2024egoexo,
  title={Ego-exo4d: Understanding skilled human activity from first-and third-person perspectives},
  author={Grauman, Kristen and Westbury, Andrew and Torresani, Lorenzo and Kitani, Kris and Malik, Jitendra and Afouras, Triantafyllos and Ashutosh, Kumar and Baiyya, Vijay and Bansal, Siddhant and Boote, Bikram and others},
  booktitle={CVPR},
  pages={19383--19400},
  year={2024}
}

@inproceedings{sener2022assembly101,
  title={Assembly101: A large-scale multi-view video dataset for understanding procedural activities},
  author={Sener, Fadime and Chatterjee, Dibyadip and Shelepov, Daniel and He, Kun and Singhania, Dipika and Wang, Robert and Yao, Angela},
  booktitle={CVPR},
  pages={21096--21106},
  year={2022}
}

@inproceedings{miech2019howto100m,
  title={Howto100m: Learning a text-video embedding by watching hundred million narrated video clips},
  author={Miech, Antoine and Zhukov, Dimitri and Alayrac, Jean-Baptiste and Tapaswi, Makarand and Laptev, Ivan and Sivic, Josef},
  booktitle={ICCV},
  pages={2630--2640},
  year={2019}
}

@inproceedings{tang2019coin,
  title={Coin: A large-scale dataset for comprehensive instructional video analysis},
  author={Tang, Yansong and Ding, Dajun and Rao, Yongming and Zheng, Yu and Zhang, Danyang and Zhao, Lili and Lu, Jiwen and Zhou, Jie},
  booktitle={CVPR},
  pages={1207--1216},
  year={2019}
}

@inproceedings{zhou2018towards_youcook2,
  title={Towards automatic learning of procedures from web instructional videos},
  author={Zhou, Luowei and Xu, Chenliang and Corso, Jason},
  booktitle={AAAI},
  volume={32},
  year={2018}
}

@inproceedings{bao2023can_WTAG,
  title={Can foundation models watch, talk and guide you step by step to make a cake?},
  author={Bao, Yuwei and Yu, Keunwoo and Zhang, Yichi and Storks, Shane and Bar-Yossef, Itamar and de la Iglesia, Alex and Su, Megan and Zheng, Xiao and Chai, Joyce},
  booktitle={EMNLP},
  pages={12325--12341},
  year={2023}
}

@article{panchal2024say_QEVD,
  title={What to say and when to say it: Live fitness coaching as a testbed for situated interaction},
  author={Panchal, Sunny and Bhattacharyya, Apratim and Berger, Guillaume and Mercier, Antoine and B{\"o}hm, Cornelius and Dietrichkeit, Florian and Pourreza, Reza and Li, Xuanlin and Madan, Pulkit and Lee, Mingu and others},
  journal={NeurIPS},
  volume={37},
  pages={75853--75882},
  year={2024}
}

@inproceedings{egoper_cvpr24,
  title={Error detection in egocentric procedural task videos},
  author={Lee, Shih-Po and Lu, Zijia and Zhang, Zekun and Hoai, Minh and Elhamifar, Ehsan},
  booktitle={CVPR},
  pages={18655--18666},
  year={2024}
}

@article{peddi2024captaincook4d,
  title={CaptainCook4D: A dataset for understanding errors in procedural activities},
  author={Peddi, Rohith and Arya, Shivvrat and Challa, Bharath and Pallapothula, Likhitha and Vyas, Akshay and Gouripeddi, Bhavya and Zhang, Qifan and Wang, Jikai and Komaragiri, Vasundhara and Ragan, Eric and others},
  journal={NeurIPS},
  volume={37},
  pages={135626--135679},
  year={2024}
}

@inproceedings{jang2019epic_tent,
  title={Epic-tent: An egocentric video dataset for camping tent assembly},
  author={Jang, Youngkyoon and Sullivan, Brian and Ludwig, Casimir and Gilchrist, Iain and Damen, Dima and Mayol-Cuevas, Walterio},
  booktitle={ICCVW},
  pages={0--0},
  year={2019}
}

@inproceedings{niu2025ovo,
  title={OVO-Bench: How Far is Your Video-LLMs from Real-World Online Video Understanding?},
  author={Niu, Junbo and Li, Yifei and Miao, Ziyang and Ge, Chunjiang and Zhou, Yuanhang and He, Qihao and Dong, Xiaoyi and Duan, Haodong and Ding, Shuangrui and Qian, Rui and others},
  booktitle={CVPR},
  pages={18902--18913},
  year={2025}
}

@article{lin2024streamingbench,
  title={Streamingbench: Assessing the gap for mllms to achieve streaming video understanding},
  author={Lin, Junming and Fang, Zheng and Chen, Chi and Wan, Zihao and Luo, Fuwen and Li, Peng and Liu, Yang and Sun, Maosong},
  journal={arXiv preprint arXiv:2411.03628},
  year={2024}
}

@inproceedings{qian2025dispider,
  title={Dispider: Enabling video llms with active real-time interaction via disentangled perception, decision, and reaction},
  author={Qian, Rui and Ding, Shuangrui and Dong, Xiaoyi and Zhang, Pan and Zang, Yuhang and Cao, Yuhang and Lin, Dahua and Wang, Jiaqi},
  booktitle={Proceedings of the Computer Vision and Pattern Recognition Conference},
  pages={24045--24055},
  year={2025}
}

@article{wang2025mmduet2,
  title={MMDuet2: Enhancing Proactive Interaction of Video MLLMs with Multi-Turn Reinforcement Learning},
  author={Wang, Yueqian and Liu, Songxiang and Wang, Disong and Xu, Nuo and Wan, Guanglu and Zhang, Huishuai and Zhao, Dongyan},
  journal={arXiv preprint arXiv:2512.06810},
  year={2025}
}

@article{xu2025streamingvlm,
  title={Streamingvlm: Real-time understanding for infinite video streams},
  author={Xu, Ruyi and Xiao, Guangxuan and Chen, Yukang and He, Liuning and Peng, Kelly and Lu, Yao and Han, Song},
  journal={arXiv preprint arXiv:2510.09608},
  year={2025}
}

@misc{seed2025vision,
  author = {{ByteDance}},
  title = {Doubao-Seed-1.6},
  year = {2025},
  howpublished = {Available at \href{https://console.volcengine.com/ark/region:ark+cn-beijing/model/detail?Id=doubao-seed-1-6}{Volcengine ARK Platform}},
}

@misc{seed2026vision,
  author = {{ByteDance}},
  title = {Doubao-Seed-1.8},
  year = {2026},
  howpublished = {Available at \href{https://console.volcengine.com/ark/region:ark+cn-beijing/model/detail?Id=doubao-seed-1-8}{Volcengine ARK Platform}},
}

@misc{gemini3pro2025,
  author = {{Google DeepMind}},
  title = {Gemini 3 Pro Model Card},
  year = {2025},
  howpublished = {Available at \href{https://storage.googleapis.com/deepmind-media/Model-Cards/Gemini-3-Pro-Model-Card.pdf}{Google DeepMind Model Cards}},
}

@misc{gemini3.1pro2025,
  author = {{Google DeepMind}},
  title = {Gemini 3.1 Pro Model Card},
  year = {2026},
  howpublished = {Available at \href{https://storage.googleapis.com/deepmind-media/Model-Cards/Gemini-3-1-Pro-Model-Card.pdf}{Google DeepMind Model Cards}},
}

@inproceedings{wang2025omnimmi,
  title={OmniMMI: A Comprehensive Multi-modal Interaction Benchmark in Streaming Video Contexts},
  author={Wang, Yuxuan and Wang, Yueqian and Chen, Bo and Wu, Tong and Zhao, Dongyan and Zheng, Zilong},
  booktitle={CVPR},
  pages={18925--18935},
  year={2025}
}

@inproceedings{carion2020end,
  title={End-to-end object detection with transformers},
  author={Carion, Nicolas and Massa, Francisco and Synnaeve, Gabriel and Usunier, Nicolas and Kirillov, Alexander and Zagoruyko, Sergey},
  booktitle={ECCV},
  pages={213--229},
  year={2020},
  organization={Springer}
}

@inproceedings{cheng2022masked,
  title={Masked-attention mask transformer for universal image segmentation},
  author={Cheng, Bowen and Misra, Ishan and Schwing, Alexander G and Kirillov, Alexander and Girdhar, Rohit},
  booktitle={CVPR},
  pages={1290--1299},
  year={2022}
}

@article{wang2025proactivevideoqa,
  title={Proactivevideoqa: A comprehensive benchmark evaluating proactive interactions in video large language models},
  author={Wang, Yueqian and Meng, Xiaojun and Wang, Yifan and Zhang, Huishuai and Zhao, Dongyan},
  journal={arXiv preprint arXiv:2507.09313},
  year={2025}
}

@article{wang2025streambridge,
  title={Streambridge: Turning your offline video large language model into a proactive streaming assistant},
  author={Wang, Haibo and Feng, Bo and Lai, Zhengfeng and Xu, Mingze and Li, Shiyu and Ge, Weifeng and Dehghan, Afshin and Cao, Meng and Huang, Ping},
  journal={arXiv preprint arXiv:2505.05467},
  year={2025}
}

@inproceedings{chen2024videollm,
  title={Videollm-online: Online video large language model for streaming video},
  author={Chen, Joya and Lv, Zhaoyang and Wu, Shiwei and Lin, Kevin Qinghong and Song, Chenan and Gao, Difei and Liu, Jia-Wei and Gao, Ziteng and Mao, Dongxing and Shou, Mike Zheng},
  booktitle={CVPR},
  pages={18407--18418},
  year={2024}
}

@article{yang2025livestar,
  title={LiveStar: Live Streaming Assistant for Real-World Online Video Understanding},
  author={Yang, Zhenyu and Zhang, Kairui and Hu, Yuhang and Wang, Bing and Qian, Shengsheng and Wen, Bin and Yang, Fan and Gao, Tingting and Dong, Weiming and Xu, Changsheng},
  journal={arXiv preprint arXiv:2511.05299},
  year={2025}
}

@inproceedings{huang2024egoexolearn,
  title={Egoexolearn: A dataset for bridging asynchronous ego-and exo-centric view of procedural activities in real world},
  author={Huang, Yifei and Chen, Guo and Xu, Jilan and Zhang, Mingfang and Yang, Lijin and Pei, Baoqi and Zhang, Hongjie and Dong, Lu and Wang, Yali and Wang, Limin and others},
  booktitle={CVPR},
  pages={22072--22086},
  year={2024}
}

@inproceedings{yang2025egolife,
  title={Egolife: Towards egocentric life assistant},
  author={Yang, Jingkang and Liu, Shuai and Guo, Hongming and Dong, Yuhao and Zhang, Xiamengwei and Zhang, Sicheng and Wang, Pengyun and Zhou, Zitang and Xie, Binzhu and Wang, Ziyue and others},
  booktitle={CVPR},
  pages={28885--28900},
  year={2025}
}

@misc{bai2025qwen3vltechnicalreport,
      title={Qwen3-VL Technical Report}, 
      author={Shuai Bai and Yuxuan Cai and Ruizhe Chen and Keqin Chen and Xionghui Chen and Zesen Cheng and Lianghao Deng and Wei Ding and Chang Gao and Chunjiang Ge and Wenbin Ge and Zhifang Guo and Qidong Huang and Jie Huang and Fei Huang and Binyuan Hui and Shutong Jiang and Zhaohai Li and Mingsheng Li and Mei Li and Kaixin Li and Zicheng Lin and Junyang Lin and Xuejing Liu and Jiawei Liu and Chenglong Liu and Yang Liu and Dayiheng Liu and Shixuan Liu and Dunjie Lu and Ruilin Luo and Chenxu Lv and Rui Men and Lingchen Meng and Xuancheng Ren and Xingzhang Ren and Sibo Song and Yuchong Sun and Jun Tang and Jianhong Tu and Jianqiang Wan and Peng Wang and Pengfei Wang and Qiuyue Wang and Yuxuan Wang and Tianbao Xie and Yiheng Xu and Haiyang Xu and Jin Xu and Zhibo Yang and Mingkun Yang and Jianxin Yang and An Yang and Bowen Yu and Fei Zhang and Hang Zhang and Xi Zhang and Bo Zheng and Humen Zhong and Jingren Zhou and Fan Zhou and Jing Zhou and Yuanzhi Zhu and Ke Zhu},
      year={2025},
      eprint={2511.21631},
      archivePrefix={arXiv},
      primaryClass={cs.CV},
      url={https://arxiv.org/abs/2511.21631}, 
}

@misc{xu2025qwen3omnitechnicalreport,
      title={Qwen3-Omni Technical Report}, 
      author={Jin Xu and Zhifang Guo and Hangrui Hu and Yunfei Chu and Xiong Wang and Jinzheng He and Yuxuan Wang and Xian Shi and Ting He and Xinfa Zhu and Yuanjun Lv and Yongqi Wang and Dake Guo and He Wang and Linhan Ma and Pei Zhang and Xinyu Zhang and Hongkun Hao and Zishan Guo and Baosong Yang and Bin Zhang and Ziyang Ma and Xipin Wei and Shuai Bai and Keqin Chen and Xuejing Liu and Peng Wang and Mingkun Yang and Dayiheng Liu and Xingzhang Ren and Bo Zheng and Rui Men and Fan Zhou and Bowen Yu and Jianxin Yang and Le Yu and Jingren Zhou and Junyang Lin},
      year={2025},
      eprint={2509.17765},
      archivePrefix={arXiv},
      primaryClass={cs.CL},
      url={https://arxiv.org/abs/2509.17765}, 
}

@article{wang2025internvl3,
  title={Internvl3. 5: Advancing open-source multimodal models in versatility, reasoning, and efficiency},
  author={Wang, Weiyun and Gao, Zhangwei and Gu, Lixin and Pu, Hengjun and Cui, Long and Wei, Xingguang and Liu, Zhaoyang and Jing, Linglin and Ye, Shenglong and Shao, Jie and others},
  journal={arXiv preprint arXiv:2508.18265},
  year={2025}
}

@article{yu2025minicpm,
  title={Minicpm-v 4.5: Cooking efficient mllms via architecture, data, and training recipe},
  author={Yu, Tianyu and Wang, Zefan and Wang, Chongyi and Huang, Fuwei and Ma, Wenshuo and He, Zhihui and Cai, Tianchi and Chen, Weize and Huang, Yuxiang and Zhao, Yuanqian and others},
  journal={arXiv preprint arXiv:2509.18154},
  year={2025}
}

@article{hurst2024gpt,
  title={Gpt-4o system card},
  author={Hurst, Aaron and Lerer, Adam and Goucher, Adam P and Perelman, Adam and Ramesh, Aditya and Clark, Aidan and Ostrow, AJ and Welihinda, Akila and Hayes, Alan and Radford, Alec and others},
  journal={arXiv preprint arXiv:2410.21276},
  year={2024}
}

@inproceedings{mathew2021docvqa,
  title={Docvqa: A dataset for vqa on document images},
  author={Mathew, Minesh and Karatzas, Dimosthenis and Jawahar, CV},
  booktitle={WACV},
  pages={2200--2209},
  year={2021}
}

@inproceedings{fu2025video,
  title={Video-mme: The first-ever comprehensive evaluation benchmark of multi-modal llms in video analysis},
  author={Fu, Chaoyou and Dai, Yuhan and Luo, Yongdong and Li, Lei and Ren, Shuhuai and Zhang, Renrui and Wang, Zihan and Zhou, Chenyu and Shen, Yunhang and Zhang, Mengdan and others},
  booktitle={CVPR},
  pages={24108--24118},
  year={2025}
}

@article{wang2024measuring,
  title={Measuring multimodal mathematical reasoning with math-vision dataset},
  author={Wang, Ke and Pan, Junting and Shi, Weikang and Lu, Zimu and Ren, Houxing and Zhou, Aojun and Zhan, Mingjie and Li, Hongsheng},
  journal={NeurIPS},
  volume={37},
  pages={95095--95169},
  year={2024}
}

@article{wu2024deepseek,
  title={Deepseek-vl2: Mixture-of-experts vision-language models for advanced multimodal understanding},
  author={Wu, Zhiyu and Chen, Xiaokang and Pan, Zizheng and Liu, Xingchao and Liu, Wen and Dai, Damai and Gao, Huazuo and Ma, Yiyang and Wu, Chengyue and Wang, Bingxuan and others},
  journal={arXiv preprint arXiv:2412.10302},
  year={2024}
}

@misc{fu2024ocrbenchv2improvedbenchmark,
    title={OCRBench v2: An Improved Benchmark for Evaluating Large Multimodal Models on Visual Text Localization and Reasoning}, 
    author={Ling Fu and Biao Yang and Zhebin Kuang and Jiajun Song and Yuzhe Li and Linghao Zhu and Qidi Luo and Xinyu Wang and Hao Lu and Mingxin Huang and Zhang Li and Guozhi Tang and Bin Shan and Chunhui Lin and Qi Liu and Binghong Wu and Hao Feng and Hao Liu and Can Huang and Jingqun Tang and Wei Chen and Lianwen Jin and Yuliang Liu and Xiang Bai},
    year={2024},
    eprint={2501.00321},
    archivePrefix={arXiv},
    primaryClass={cs.CV}
}

@inproceedings{yu2019activitynet,
  title={Activitynet-qa: A dataset for understanding complex web videos via question answering},
  author={Yu, Zhou and Xu, Dejing and Yu, Jun and Yu, Ting and Zhao, Zhou and Zhuang, Yueting and Tao, Dacheng},
  booktitle={AAAI},
  pages={9127--9134},
  year={2019}
}

@inproceedings{zhang2024mathverse,
  title={Mathverse: Does your multi-modal llm truly see the diagrams in visual math problems?},
  author={Zhang, Renrui and Jiang, Dongzhi and Zhang, Yichi and Lin, Haokun and Guo, Ziyu and Qiu, Pengshuo and Zhou, Aojun and Lu, Pan and Chang, Kai-Wei and Qiao, Yu and others},
  booktitle={ECCV},
  pages={169--186},
  year={2024},
  organization={Springer}
}

@article{zheng2023judging,
  title={Judging llm-as-a-judge with mt-bench and chatbot arena},
  author={Zheng, Lianmin and Chiang, Wei-Lin and Sheng, Ying and Zhuang, Siyuan and Wu, Zhanghao and Zhuang, Yonghao and Lin, Zi and Li, Zhuohan and Li, Dacheng and Xing, Eric and others},
  journal={NeurIPS},
  volume={36},
  pages={46595--46623},
  year={2023}
}

@misc{qwen35blog,
    title = {Qwen3.5: Accelerating Productivity with Native Multimodal Agents},
    url = {https://qwen.ai/blog?id=qwen3.5},
    author = {Qwen Team},
    month = {February},
    year = {2026}
}

@article{singh2025openai,
  title={Openai gpt-5 system card},
  author={Singh, Aaditya and Fry, Adam and Perelman, Adam and Tart, Adam and Ganesh, Adi and El-Kishky, Ahmed and McLaughlin, Aidan and Low, Aiden and Ostrow, AJ and Ananthram, Akhila and others},
  journal={arXiv preprint arXiv:2601.03267},
  year={2025}
}

@inproceedings{ben2021ikea,
  title={The ikea asm dataset: Understanding people assembling furniture through actions, objects and pose},
  author={Ben-Shabat, Yizhak and Yu, Xin and Saleh, Fatemeh and Campbell, Dylan and Rodriguez-Opazo, Cristian and Li, Hongdong and Gould, Stephen},
  booktitle={WACV},
  pages={847--859},
  year={2021}
}

@article{lu2026aura,
  title={Aura: Always-on understanding and real-time assistance via video streams},
  author={Lu, Xudong and Bo, Yang and Chen, Jinpeng and Li, Shuhan and Guo, Xintong and Guan, Huankang and Liu, Fang and Xu, Dunyuan and Sun, Peiwen and Sun, Heyang and others},
  journal={arXiv preprint arXiv:2604.04184},
  year={2026}
}

@inproceedings{lu2026phostream,
  title={PhoStream: Benchmarking Real-World Streaming for Omnimodal Assistants in Mobile Scenarios},
  author={Lu, Xudong and Guan, Huankang and Bo, Yang and Chen, Jinpeng and Guo, Xintong and Li, Shuhan and Liu, Fang and Sun, Peiwen and Li, Xueying and Zhang, Wei and others},
  booktitle={ICML},
  year={2026}
}

@inproceedings{sun2025spacevista,
  title={Spacevista: All-scale visual spatial reasoning from mm to km},
  author={Sun, Peiwen and Lang, Shiqiang and Wu, Dongming and Ding, Yi and Feng, Kaituo and Liu, Huadai and Ye, Zhen and Liu, Rui and Liu, Yun-Hui and Wang, Jianan and others},
  booktitle={ICML},
  year={2026}
}

@inproceedings{sun2026x,
title={X-Stream: Exploring MLLMs as Multiplexers for Multi-Stream Understanding},
author={Sun, Peiwen and Lu, Xudong and Liu, Huadai and Bo, Yang and Wu, Dongming and Guan, Huankang and Cai, Minghong and Chen, Jinpeng and Guo, Xintong and Li, Shuhan and others},
booktitle={ECCV},
year={2026}
}

@inproceedings{ding2025streammind,
  title={Streammind: Unlocking full frame rate streaming video dialogue through event-gated cognition},
  author={Ding, Xin and Wu, Hao and Yang, Yifan and Jiang, Shiqi and Zhang, Qianxi and Bai, Donglin and Chen, Zhibo and Cao, Ting},
  booktitle={ICCV},
  pages={13448--13459},
  year={2025}
}

@inproceedings{lin2026hippomm,
  title={Hippomm: Hippocampal-inspired multimodal memory for long audiovisual event understanding},
  author={Lin, Yueqian and Zhang, Jingyang and Wang, Qinsi and Ye, Hancheng and Fu, Yuzhe and Liu, Yudong and Li, Hai Helen and Chen, Yiran},
  booktitle={CVPR},
  pages={5968--5977},
  year={2026}
}

@inproceedings{long2025seeing_m3agent,
  title={Seeing, listening, remembering, and reasoning: A multimodal agent with long-term memory},
  author={Long, Lin and He, Yichen and Ye, Wentao and Pan, Yiyuan and Lin, Yuan and Li, Hang and Zhao, Junbo and Li, Wei},
  booktitle={ICLR},
  year={2026}
}

@article{yang2025svbench,
  title={Svbench: A benchmark with temporal multi-turn dialogues for streaming video understanding},
  author={Yang, Zhenyu and Hu, Yuhang and Du, Zemin and Xue, Dizhan and Qian, Shengsheng and Wu, Jiahong and Yang, Fan and Dong, Weiming and Xu, Changsheng},
  journal={arXiv preprint arXiv:2502.10810},
  year={2025}
}

@inproceedings{liu2026gensplat,
  title={GenSplat: Bridging the Generalization Gap in 3DGS Language Comprehension},
  author={Liu, Fang and Liu, Yuhao and Xu, Ke and Hancke, Gerhard Petrus and Lau, Rynson WH},
  booktitle={CVPR},
  pages={5221--5231},
  year={2026}
}

@inproceedings{liu2025language,
  title={Language-guided salient object ranking},
  author={Liu, Fang and Liu, Yuhao and Xu, Ke and Ye, Shuquan and Hancke, Gerhard Petrus and Lau, Rynson WH},
  booktitle={CVPR},
  pages={29803--29813},
  year={2025}
}

@inproceedings{liu2023referring,
  title={Referring image segmentation using text supervision},
  author={Liu, Fang and Liu, Yuhao and Kong, Yuqiu and Xu, Ke and Zhang, Lihe and Yin, Baocai and Hancke, Gerhard and Lau, Rynson},
  booktitle={ICCV},
  pages={22124--22134},
  year={2023}
}

\end{document}